\documentclass[sigconf]{acmart}

\usepackage{booktabs} 
\usepackage{graphicx} 
\usepackage{multirow}
\usepackage{balance}
\usepackage{lipsum}

\newcommand\blfootnote[1]{%
  \begingroup
  \renewcommand\thefootnote{}\footnote{#1}%
  \addtocounter{footnote}{-1}%
  \endgroup
}

\setcopyright{rightsretained}
\acmConference[IDM 2017]{CIKM Workshop on Interpretable Data Mining}{November 2017}{Singapore} 

\copyrightyear{2017}

\renewcommand\footnotetextcopyrightpermission[1]{} 

\begin{document}

\title{Interpretable Feature Recommendation for Signal Analytics}
\author{\textbf{Snehasis Banerjee}}
\affiliation{%
  \institution{TCS Research \& Innovation}
  \streetaddress{Tata Consultancy Services}
  \city{Kolkata} 
  \state{West Bengal} 
  \country{India} 
  \postcode{700160}
}
\email{snehasis.banerjee@tcs.com}

\author{\textbf{Tanushyam Chattopadhyay}}
\affiliation{%
  \institution{TCS Research \& Innovation}
  \streetaddress{Tata Consultancy Services}
  \city{Kolkata} 
  \state{West Bengal} 
  \country{India} 
  \postcode{700160}
}
\email{t.chattopadhyay@tcs.com}

\author{\textbf{Ayan Mukherjee}}
\affiliation{%
  \institution{TCS Research \& Innovation}
  \streetaddress{Tata Consultancy Services}
  \city{Kolkata} 
  \state{West Bengal} 
  \country{India} 
  \postcode{700160}
}
\email{ayan.m4@tcs.com}


\begin{abstract}
This paper presents an automated approach for interpretable feature recommendation for solving signal data analytics problems. The method has been tested by performing experiments on datasets in the domain of prognostics where interpretation of features is considered very important. The proposed approach is based on Wide Learning architecture and provides means for interpretation of the recommended features. It is to be noted that such an interpretation is not available with feature learning approaches like Deep Learning (such as Convolutional Neural Network) or feature transformation approaches like Principal Component Analysis. Results show that the feature recommendation and interpretation techniques are quite effective for the problems at hand in terms of performance and drastic reduction in time to develop a solution. It is further shown by an example, how this human-in-loop interpretation system can be used as a prescriptive system. \blfootnote{Copyright retained by the authors, 2017.}
\end{abstract}



\maketitle

\section{Introduction}
Development of a sensor data based descriptive and prescriptive system involves machine learning tasks like classification and regression. Any such system development requires the involvement of different stake-holders like:\\
\textit{Domain expert}: who understand the problem domain and can make sense of features of a model for causality analysis, ex. a mechanical engineer of a machine plant in case of machine prognostics\\
\textit{Signal processing (SP) expert}: who can suggest suitable signal processing algorithms (such as spectrogram) and their corresponding tuning parameters (such as spectrum type and window overlap)\\
\textit{Machine Learning (ML) expert}: who can perform data analysis and design the models for a ML task such as classification or regression\\
\textit{Coder or developer}: who can construct a deployable solution to be used by end users, after other stakeholders have shared inputs.

Now, the problem of developing such a system is that each of the stake holders speaks their own language and terms. The typical work-flow steps for such a sensor data analytics task is as follows:\\
1. Domain Expert explains the goal of the problem pertaining to the use case and application to the SP and ML resource persons.\\
2. SP expert provides a list of algorithms that can be used as features (data transforms to make data easy for analysis) for given problem.\\
3. ML expert recommends the optimal feature set based on analysis of the available dataset and her/his knowledge of similar problems.\\
4. SP expert tunes the parameters of those algorithms (such as window size, n-point for a Fast Fourier Transform algorithm), and the ML expert tunes the (hyper) parameters to derive a solution model.\\
5. Recommended feature set is presented to domain expert for validation and verification, to check if extracted features are sound.\\
6. If current features are meaningless, which is often the case, the steps 2-5 are repeated in iteration with a change in approach and taking into account the domain expert's feedback.\\
7. Final system with finalized model is deployed by developer.\\
In a related survey by \cite{banerjee2017survey}, it was found that the most difficult task in the above work-flow is task 2 and 3, namely feature engineering (a combination of feature listing / extraction and feature selection). Step 5 (validation of features by domain experts) is difficult in a Deep Learning based approach as features obtained are not interpretable for 1-D sensor signals. The paper presents how to interpret the recommended features for machine prognostics and activity monitoring by using a modified \textit{Wide Learning} (\cite{banerjee2016towards}) approach that was earlier found useful for health-care domain \cite{banerjee2016time}.\\
The main contributions are: a) Description of the method with focus on interpretation b) Comparative results of the \textit{Wide} method with standard methods c) Illustration of interpretable features obtained.

\section{Method Description}

\begin{figure}[t]
\centering
\includegraphics[width=3.5in]{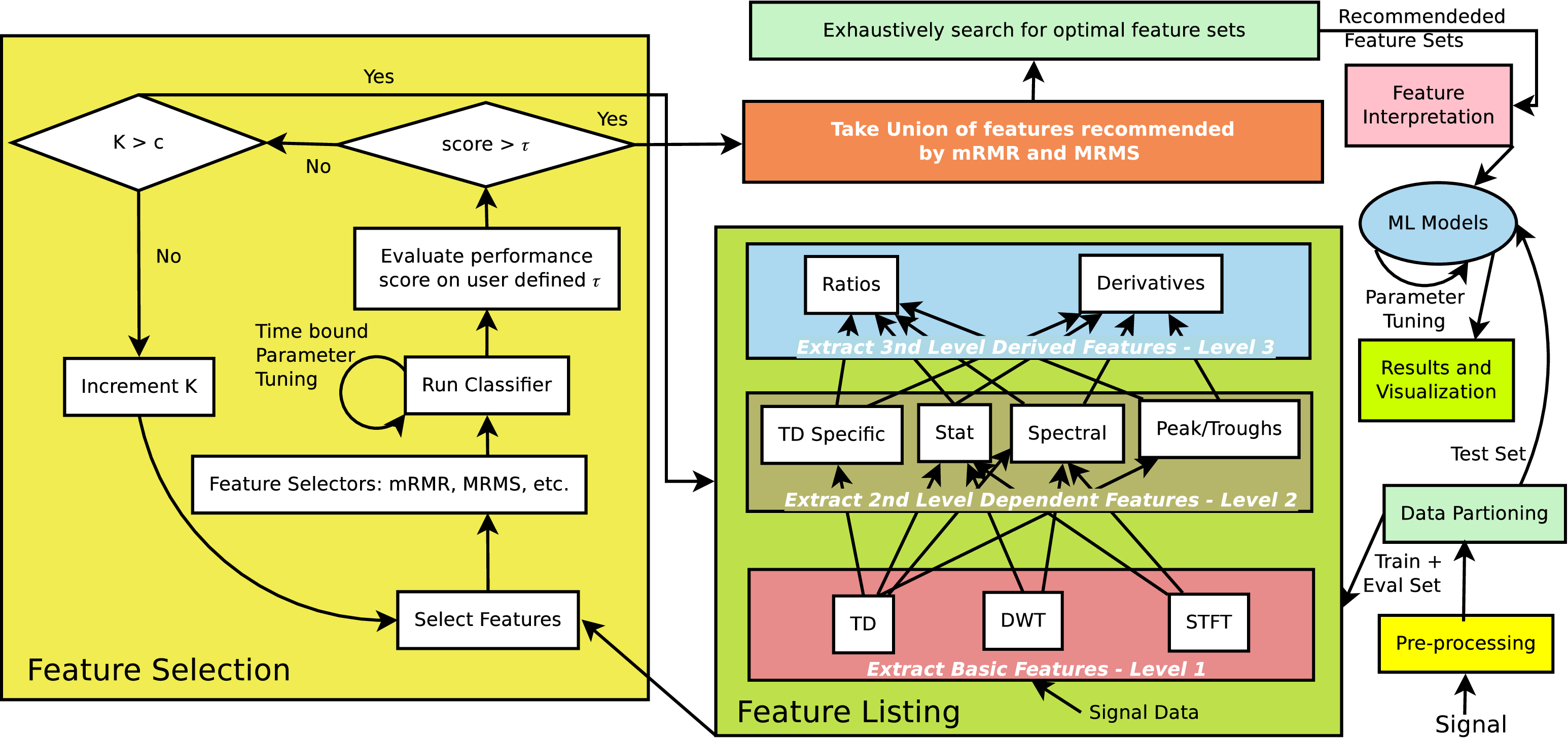}
\caption{Wide Method for Feature Engineering}
\label{fig:FeatureEngineering}
\end{figure}

The \textit{Wide Learning} system as shown in Fig. 1 accepts a set of annotated input signal data. The signal data is obtained after a) standard pre-processing steps are executed and final outcome is a standard matrix format with labels b) data is partitioned into Train and Test in multiple folds (usually 5). The system automatically determines the number of folds
depending on number of data instances available. The partitioning of data takes place following Train-Eval-Test principles (\cite{Andrew-2017}) in folds, with each partition retaining data characteristics based on data clustering. The number of clusters were determined based on cluster quality metric namely Silhouette Coefficient. The data partitions were composed of proportionate distribution (as per number of folds) of cluster members among Train-Eval-Test sets. The performance (say accuracy) is reported on the hidden Test Set, while the rest is used for feature recommendation. The Train data is passed to extract the features at various levels of feature extraction. The `Eval' Set is used for classifier-in loop evaluation (wrapper method of feature selection) on obtained features derived from the Train set. The Classifiers used are an ensemble of Random Forest and linear and Gaussian kernels for Support Vector Machine (SVM) with time bounded parameter tuning. The intuition is that even using under-tuned models, good features reveal themselves. 

Basic features reported in literature of sensor data analytics can be mainly classified in three types: (i) Time Domain features (TD) (ii) Fourier transformation based features (STFT) (iii) Discrete Wavelet transformation based features (DWT). So, at Level 1, basic features are extracted and passed on to Feature Selection Module. DWT requires input of a mother wavelet type\footnote{Various Wavelet Family Listing: http://in.mathworks.com/help/wavelet/gs/introduction-to-the-wavelet-families.html} as one of its parameters, but automated mother wavelet identification is a challenging problem. The appropriate mother wavelet for carrying out the Wavelet Transform is selected by comparing the input signal with a library of mother wavelets in terms of having maximum energy to entropy ratio. As the purpose of a feature is distinguish between two groups, so an alternative less error-prone distance base approach is also applied. Here, each mother wavelet's energy-entropy ratio is ranked and the one that has maximum distance to the a set of training classes are added as a feature.  In level 2, spectral, statistical, time domain based and peak-trough features are extracted. Level 3 includes different ratios and derivatives of the level 2 features. Feature subsets are selected by iteratively applying a combination of two powerful feature selection techniques in the wrapper approach, namely mRMR (\cite{peng2005feature}) and MRMS (\cite{maji2012fuzzy}). They cover different aspects of feature selection. For instance, mRMR is classifier independent where as MRMS is effective to reduce real valued noisy features which are likely to occur in sensor data. The system is open to add more feature selectors as per need. The system finds 2 feature sets of cardinality `k' for a particular performance metric (such as accuracy, sensitivity, specificity): a) Fe1 - that produces the highest metric in any fold of cross-validation b) Fe2 - that is most consistent and performs well across all folds. The above step of feature selection is done hierarchically - if layer 1 does not produce expected results set by a user defined pre-set threshold $\tau$ or maximum possible goodness value of a selected metric (say 1.0 on a scale of 0 to 1 for a metric accuracy), then layer 2 (higher level features) is invoked, and so on. `c' is a regularizer for `k' and is dependent proportionately on the hardware capabilities of the experimentation system. Post feature selection, an exhaustive search is done on the finalized `k' features to find the ideal feature combination (best among $2^k-1$ subsets) for the task. It has been shown in literature that without applying brute-force, apt feature combination cannot be arrived with certainty. This selected feature recommendation set is used for modeling using standard classifiers like Artificial Neural Network (ANN), SVM, Random Forest post parameter tuning to derive results on the hidden Test Set. 

Despite the advantages, the system has the following limitations:\\
a) Exponential time to find optimal feature set to try out all combinations (for k $>$ 30, it becomes impractical on standard hardware).\\
b) Currently performance optimization is restricted to 1 metric (say sensitivity), but often use cases demand multi-metric optimization.\\
c) Restriction to union of mRMR and MRMS as feature selection, other methods needs evaluation and integration.\\
d) Non-exhaustiveness in the super set of features for a task in the feature listing database, unknown features will be missed.\\
e) Data geometry analysis and data imbalance handling needs to be addressed to obtain better performance and yield a robust model.\\
f) Feature recommendation is assumed as classifier independent.

\subsection{Feature Interpretation Module}

Tables \ref{2fet} and \ref{3fet} show some of the sample feature sets obtained for the classification task in dataset D1 (Nasa Bearing). It can be seen that recommended features differ based on specified window size. The window size plays a major role which is usually supplied by the domain expert (for dataset D1 ideal window size is 1 second as per literature). This listing of features along with ranges of values obtained for the feature type aids the domain experts to map the obtained feature values to the physical world and the problem domain, so that deeper insights can be gained. 

In general, any feature set recommendation framework would recommend only the corresponding indices of the relevant features. Such feature identification mechanism is sufficient to trace back the recommended features from the generated feature pool. However, such a practice do not leave any room for further refinement of the recommendation through incorporation of domain expert's opinion. Also, when dealing with windowed processing, often the same features of different windows can get reported. So there needs to be means to identify features in different windows and compress them together instead of multiple window-wise reporting in cases of non-time variation dependent features. To address this issue, the proposed framework consists of a feature interpretation module. This module accepts the recommended feature indices as input and returns any granular information that can be obtained by analyzing its step-by-step genesis process across windows of data processing. While feature values were derived to form input derived feature pool, a mapping table is iteratively maintained that stores the details of the steps through which each indexed feature value is being generated. The steps of each indexed value generation would typically include information regarding domain of transformation, transformation technique, location of the feature value in the transformed vector, etc. This is in contrast to a hard-coded repository of feature names tagged to unique identifiers, so that new feature extracting modules can be added and the meta-data update happens at the time of component plug-in. A format for feature extraction algorithm entry in database is maintained, that include algorithm description, value ranges which can aid in interpretation later.  Another feature is that domain experts can add weights to those features which seem to have a physical world connection, so that related feature space can be explored. As an example, if domain experts tag spectral features as relevant, more parameter tuning will be carried out on a variety of spectral features. Integration with knowledge-bases (OWL based ontologies and probabilistic rules) \cite{banerjee2014semantic} along with window based stream reasoning \cite{banerjee2013windowing} \cite{mukherjee2013towards} \cite{mukherjee2013system} is planned in future.

\begin{figure}[t]
\centering
\includegraphics[width=3.5in]{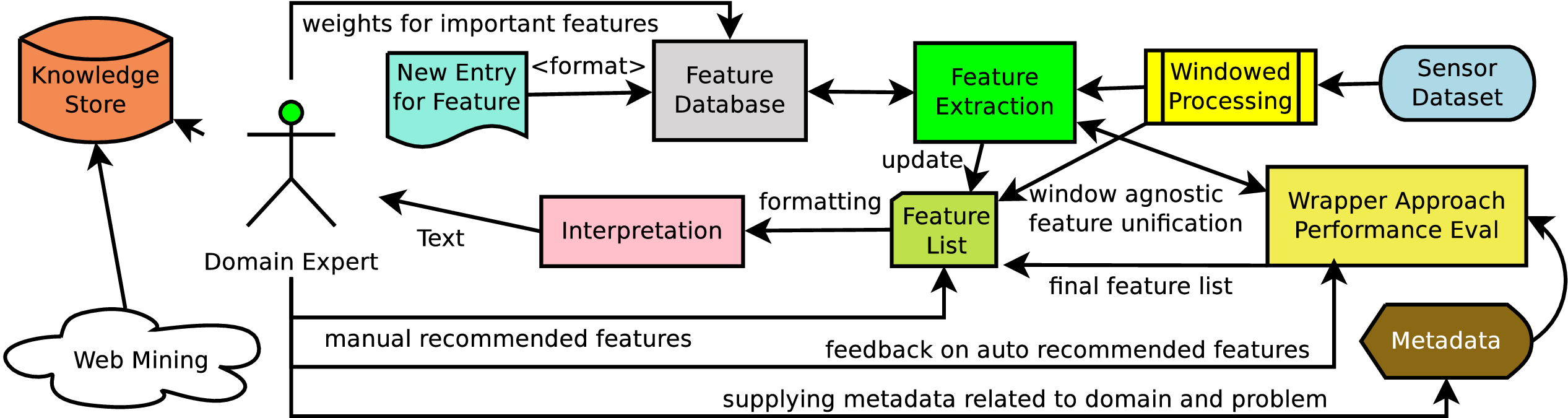}
\caption{Feature Interpretation module}
\label{fig:FeatureEngineering}
\end{figure}
\begin{table}[t]
		\caption{Recommended features for D1, window size = 0.5 sec}
		\begin{center}
			\begin{tabular}{|p{0.05\linewidth}|p{0.25\linewidth}|p{0.45\linewidth}|}
				\hline
				Sl. & \multicolumn{2}{p{0.7\linewidth}|}{\hspace{5em}Feature description} \\
				\hline
				\multirow{5}{*}{1} & \multirow{5}{*}{STFT}& Frequency: 1851.1851 Hz\\
               & & Frequency: 1853.1853 Hz \\
			   & & Frequency: 1153.1153 Hz \\
               & & Frequency: 1837.1837 Hz \\
               & & Frequency: 1845.1845 Hz \\
				\hline
				2      &\multicolumn{2}{p{0.7\linewidth}|}{Difference of standard deviation values of windowed discrete wavelet transform (DWT) coefficients}                        \\
				\hline
				3      & \multicolumn{2}{p{0.7\linewidth}|}{Standard deviation of STFT coefficients}  \\
				\hline
				4      & \multicolumn{2}{p{0.7\linewidth}|}{DWT Frequency: (harmonic) 14.4991 Hz}  \\
				\hline			
			\end{tabular}
			
		\end{center}
		\label{2fet}
	\end{table}
	\begin{table}[t]
		\caption{Recommended features for D1, window size = 1 sec}
		\begin{center}
			\begin{tabular}{|p{0.05\linewidth}|p{0.25\linewidth}|p{0.45\linewidth}|}
				\hline
				Sl. & \multicolumn{2}{p{0.7\linewidth}|}{\hspace{5em}Feature description} \\
				\hline
				\multirow{4}{*}{1} & \multirow{4}{*}{STFT} & Frequency: 1613.5807 Hz \\
				& & Frequency: 1829.5915 Hz  \\
				& & Frequency: 1830.5915 Hz \\
				& & Frequency: 1837.5919 Hz \\
				\hline
				2      &\multicolumn{2}{p{0.7\linewidth}|}{Kurtosis of DWT coefficients}                        \\
				\hline
				3      & \multicolumn{2}{p{0.7\linewidth}|}{Standard deviation of DWT coefficients}  \\
				\hline
				4      &\multicolumn{2}{p{0.7\linewidth}|}{Standard deviation of STFT coefficients}                        \\
				\hline
				5      & \multicolumn{2}{p{0.7\linewidth}|}{Zero crossing of DWT coefficients}  \\
				\hline
				6      & \multicolumn{2}{p{0.7\linewidth}|}{DWT Frequency: (harmonic) 14.3701 Hz}  \\
				\hline
			\end{tabular}
		\end{center}
		\label{3fet}
	\end{table}
\section{Experiments}
\subsection{Datasets}
\label{Datasets}
The experiment is performed on two popular and open 1-dimensional sensor signal data sets, the specification being tabulated in table 3 and described as follows:

(i) D1 and D2: NASA Bearing\footnote{NASA Bearing Set 3 at https://ti.arc.nasa.gov/tech/dash/pcoe/ prognostic-data-repository/publications/\#bearing} data set contains 4 bearing data instances each having 984 records, while the first bearing fails after 700th record among the total 984 recorded readings. The last two readings are not considered due to presence of missing values. So, we get 282 `bad bearing' (class 0) records as ground truth for a class, while the rest 700 of the first bearing and 982 values each from rest 3 bearings that do not fail form the `good bearing' class 1. To handle data unbalancing and see its effects, we have created two data-sets: D1: that contains the full dataset instances, D2: that contains a randomly selected small subset of the `good bearing' instances along with all the `bad bearing' instances. We have restricted to binary classification tasks to get comparable results.

\begin{table*}[h]
\caption{Description of data sets used for experiments}
\label{Table:comparison}
\centering
\begin{tabular}{|l|c|c|c|c|c|c|}
\hline
Datasets (D) &Total No. of    &Class-0 No. of  &Class-1 No. of     &No. of     &Sampling\\
         &Instances  &Instances         &Instances      &Samples      &Rate (Hz)\\
\hline
D1: NASA All &3932     &282         &3650             &20480     &20,000\\
\hline
D2: NASA Subset &647     &282         &365             &20480     &20,000\\
\hline
D3: Mobifall &258     &132         &126             &230     &50\\
\hline
\end{tabular}
\end{table*}

(ii) D3: Mobifall\footnote{Mobifall Challenge Dataset at http://www.bmi.teicrete.gr/index.php/research/mobiact} data set is a popular fall detection data-set created by volunteers aged 22-47 years. Although the data-set contains various levels of activities, however the data-set was partitioned into `fall' (class 0) and `not fall' (class 1), in order to restrict to binary classification task.

\begin{table*}[t]
\caption{Comparison in terms of accuracy (PCA, MLP, CNN, LSTM, manual SoA, WIDE method)}
\label{Table:comparison}
\centering
\begin{tabular}{|l|c|c|c|c|c|c|c|c|c|}
\hline
Datasets (D) & PCA & MLP  & CNN & LSTM$^\$$ & MLP$^*$  & LSTM$^*$ & manual SoA & WIDE\\
\hline
D1. Nasa All &0.94  &0.93   &  0.93   &  0.93    &  0.96     &  0.97    &  0.99       &  1.0\\
\hline
D2. Nasa Subset  &0.52  &0.5   &  0.5   &  0.5    &  0.55     &  0.56    &  0.99       &  1.0\\
\hline
D3. Mobifall  &0.51 &0.44   &  0.44   &  0.44    &  0.44      &  0.44    &  0.99       &  0.98\\
\hline

Approx. Effort   & D1. 2   & D1. 3 & D1. 5 & D1. 6  & D1. 4  & D1. 7  & D1. 30 & D1. 1\\
in person-days & D2. 2   & D2. 3 & D2. 5 & D2. 8  & D2. 4  & D2. 9 & D2. 30 & D2. 1\\
unit for the task   & D3. 1  & D3. 4 & D3. 7 & D3. 9  & D3. 4.2 & D3. 9.2 & D3. 90 & D3. 0.2\\
\hline
Interpretable &No & No & No & No & Yes & Yes & Yes & Yes \\
\hline
\end{tabular}
\begin{tabular}{l}
\small $^\$$ output of CNN layers are fed to LSTM ; $^*$ performance measured on features extracted by the \textit{Wide} method
\end{tabular}
\end{table*}

\subsection{Experimental Setup}
Deep Learning based experiments has been carried out using Theano\footnote{Theano v. 0.8.2, http://deeplearning.net/software/theano} on a 8-core Intel 2.66 GHz machine having nVidia GTX 1080 GPU. Multi-layer Perceptron (MLP), Convolutional Neural Network (CNN) and Long-Short Term Memory (LSTM) based Recurrent Neural Network were configured following standard rules of thumbs and principles to obtain results on the 3 datasets with grid search based hyper parameter optimization. Principal component analysis (PCA) is a statistical procedure that uses an orthogonal transformation to derive principal components representative of the features under consideration. Experiments has been carried out on above datasets with both linear and Gaussian SVM kernels with varying number of principal components obtained post application of PCA.  

\subsection{Results and Analysis}
Table \ref{Table:comparison} lists the obtained result for a dataset along with the corresponding effort for each of PCA (with SVM as classifier), MLP, CNN, LSTM, state-of-art (SoA) and proposed Wide method. It shows that PCA based methods (where features are projections and not interpretable) are outperformed by \textit{Wide} method. Deep Learning (DL) approaches were applied on both raw data as well as features recommended by proposed method. It is seen that DL based techniques fail when compared to SoA and the proposed Wide Learning method, probably because of less data instances. The two major problems with DL is (a) It needs a lot of data for training which is often not available for 1-D sensor signals. Moreover, the data availability is skewed where mostly data of `good' class is available, with trace amounts of `bad' class (say healthy and failing machine parts) (b) There is no way to interpret the features for causal analysis. It was observed that DL techniques classify all the test instances into one class that can be found by calculating the ratio between classes of table 3 (apart from confusion matrix) for NASA bearing dataset D1 and D2. Another notable observation is that, in no instance, has classification performance on recommended features fallen in comparison with automated feature learning. The performance for Mobifall dataset is not at par in case of DL that can be attributed to the low number of input vectors for training the deep models. Hence, the proposed Wide Learning approach was found to be effective for the above cases with huge reduction of development time and at par performance.

\section{Physical interpretation}

Traditionally feature selection method is a manual effort where a domain expert identifies some features using her/his domain expertise and experience; and then plot them for various class labels to conclude whether the features are relevant or not for a problem. In line with that the NASA Bearing dataset is selected here for interpretation analysis. Similar interpretation were also found in the other data set. The automated feature recommendation method predicted features at 14 Hz (DWT feature) harmonic space of the fundamental frequencies of the bearings rotating elements as reported below. Therefore the recommended features can be mapped to the physical world elements for further introspection and analysis by the in-loop domain expert. The bearing physics as per literature suggests fundamental frequencies as:\\
a) Outer Race Frequency = 236.4 Hz\\
b) Inner Race  Frequency = 296.9 Hz\\
c) Rolling Element Frequency = 279.8 Hz\\
d) Shaft Frequency = 33.33 Hz\\
e) Bearing Cage Frequency = 14.7 Hz.\\
In this case, it can be predicted that bearing fault may arise because of all possible reasons other than the problem in Shaft frequency (features do not reveal that frequency as a differentiator), where as Bearing Cage frequency seems to be the most causally related to failure. Hence, the reasons of failure can be suggested to the manufacturer by physical interpretation of the recommended features, and its mapping to the physical world for future defect prevention.

\section{Conclusion}
This paper presented a method to recommend features based on a Wide Learning technique that the domain experts can interpret. In case of NASA bearing data-set (for machine prognostics), this interpretation was found to help them to analyze the probable cause of failure. The experimental results obtained on two typical datasets by comparing standard methods (state of art of the problem, deep learning approaches, PCA) with the proposed approach proved its effectiveness in terms of both performance as well as drastic reduction in time to develop a prototype solution. The current focus of the work was 1-D signal, future work will explore similar approaches for 2-D (image) and 3-D (video) signal processing.

\bibliographystyle{ACM-Reference-Format}
\bibliography{cikmW}

\balance

\end{document}